\documentclass[runningheads]{llncs}
\usepackage{graphicx}

\usepackage{tikz}
\usepackage{comment}
\usepackage{amsmath,amssymb} 
\usepackage{color}
\usepackage{booktabs}
\usepackage{multirow}
\usepackage{bm}

\usepackage[accsupp]{axessibility}  

\begin{document}
\pagestyle{headings}
\mainmatter
\def\ECCVSubNumber{1958}  


\title{Language Matters: A Weakly Supervised Vision-Language Pre-training Approach for Scene Text Detection and Spotting}


\titlerunning{oCLIP}
%
\author{Chuhui Xue \inst{1,2}\and
Wenqing Zhang \inst{2} \and
Yu Hao\inst{2} \and \\
Shijian Lu \inst{1} \and
Philip Torr \inst{3}\and
Song Bai \inst{2}}
\authorrunning{Xue et al.}
\institute{Nanyang Technological University \and ByteDance Inc.\and University of Oxford}
\maketitle

\begin{abstract}


Recently, Vision-Language Pre-training (VLP) techniques have greatly benefited various vision-language tasks by jointly learning visual and textual representations, which intuitively helps in Optical Character Recognition (OCR) tasks due to the rich visual and textual information in scene text images. However, these methods cannot well cope with OCR tasks because of the difficulty in both instance-level text encoding and image-text pair acquisition (i.e. images and captured texts in them). This paper presents a weakly supervised pre-training method, oCLIP, which can acquire effective scene text representations by jointly learning and aligning visual and textual information. Our network consists of an image encoder and a character-aware text encoder that extract visual and textual features, respectively, as well as a visual-textual decoder that models the interaction among textual and visual features for learning effective scene text representations. With the learning of textual features, the pre-trained model can attend texts in images well with character awareness. Besides, these designs enable the learning from weakly annotated texts (i.e. partial texts in images without text bounding boxes) which mitigates the data annotation constraint greatly. Experiments over the weakly annotated images in ICDAR2019-LSVT show that our pre-trained model improves F-score by +2.5\% and +4.8\% while transferring its weights to other text detection and spotting networks, respectively. In addition, the proposed method outperforms existing pre-training techniques consistently across multiple public datasets (e.g., +3.2\% and +1.3\% for Total-Text and CTW1500).

\end{abstract}

\section{Introduction}

\begin{figure}[t!]
  \centering
  \includegraphics[width=\linewidth]{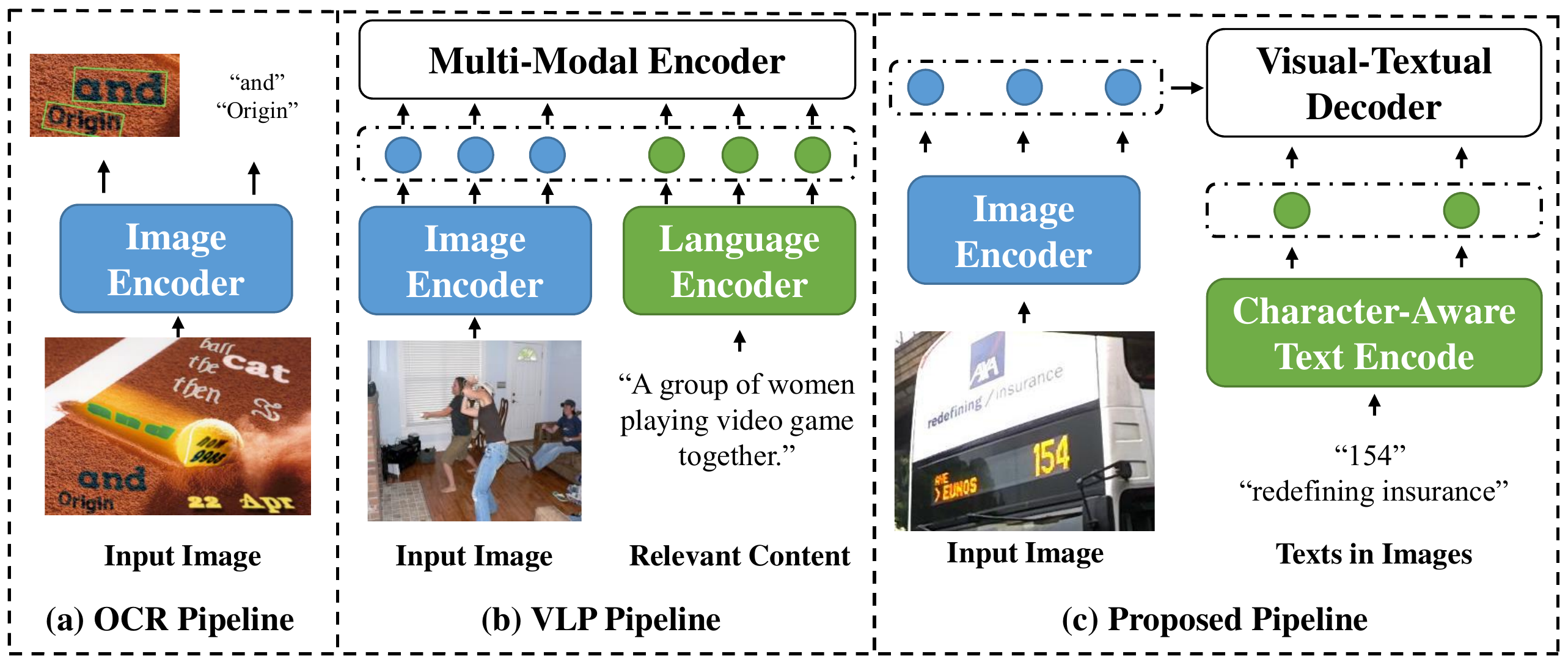}
  \caption{\textbf{Illustration of general Optical Character Recognition (OCR), Vision-Language Pre-training (VLP) pipeline, and the proposed pipeline:} General OCR pipelines focus only on visual features from images. In addition, general VLP models extract image and language features from input images and corresponding sentence-level text, and model the interaction among all visual and textual features through a multi-modal encoder. Differently, the proposed model extracts instance-level textual features from texts instances in images. It models the interactions between each text instance and its extracted image features which can be trained with weak supervision only (i.e. partial texts in images without text bounding boxes). Our pre-trained model weights can be directly transferred to various scene text detectors and spotters with significant performance improvement.
  } 
  \label{fig:intro}
\end{figure}

Optical Character Recognition (OCR) (including scene text detection, recognition, and spotting) has attracted increasing interests in recent years in both computer vision and deep learning research communities due to its wide range of applications in multilingual translation, autonomous driving, etc. Most of the existing OCR techniques follow general computer vision pipelines that first extract visual features from the input image and and then perform feature regression or classification for text detection or recognition, as shown in Fig. \ref{fig:intro} (a). However, we human usually read texts by utilizing not only the visual features of each text but also our linguistic knowledge in our memory. For example, we usually locate and read texts faster and more easily with the knowledge of the corresponding text language. Therefore, both visual and textual information are useful to robust reading of texts from natural scene images.


Recently, joint learning visual and textual representations has been studied in many Vision-Language Pre-training (VLP) techniques \cite{radford2021learning,xue2021probing,chen2020uniter}, and it greatly promotes various Vision-Language (VL) tasks such as Visual Question Answering (VQA), Image-Text Retrieval, etc. As a language-related task, OCR can intuitively benefit from these VLP techniques. However, most existing VLP methods usually suffer from two typical constraints while being applied to OCR tasks.
\textbf{(1)} Each image in VL tasks is usually associated with one sentence or paragraph where words or phrases (i.e. tokens) are arranged in reading orders. Instead, an image in OCR tasks often contains many text instances each of which consists of one or multiple tokens. The tokens within one text instance are often closely related to each other (e.g. `redefining' and `insurance' in Fig. \ref{fig:intro}(c)) while those from different text instances are completely irrelevant (e.g. `insurance' and `154' in Fig. \ref{fig:intro}(c)). This makes it difficult to encode the textual information in a general sequential way.
\textbf{(2)} Most VLP models learn from image-text pairs in which images and texts are correlated with each other at content-level  (e.g. images and captions) as illustrated in Fig. \ref{fig:intro}(b). These content-relevant image-text pairs can be easily obtained from web, social media, etc., which has been proven to be effective for various VL tasks \cite{radford2021learning}. In contrast, OCR tasks aim to detect and recognize text instances that appear in images as shown in Fig. \ref{fig:intro}(c). The image-text pairs (i.e. images and texts in them) are more difficult to obtain as compared to VL tasks, requiring expensive and inefficient annotations.

We present an \textbf{\underline{O}}CR \textbf{\underline{C}}ontrastive \textbf{\underline{L}}anguage-\textbf{\underline{I}}mage \textbf{\underline{P}}re-training (oCLIP) technique that exploits textual information for learning effective visual text representations for better scene text detection and spotting. Different from the text encoder in the existing VLP methods \cite{radford2021learning}, we design a character-aware text encoder as illustrated in Fig. \ref{fig:intro}(c). It extracts language features by encoding textual information from the sequence of characters in each text instance without considering the relations among irrelevant text instances. In addition, we introduce a visual-textual decoder that models the relations between the input image and each labelled text instance only instead of all captured texts in the input image. With the two designs, our network can learn effective visual text representations from weakly-annotated data (i.e. partial text instances in images without text bounding boxes) which greatly mitigates the data acquisition challenge and enables exploitation of large amounts of weakly-annotated images.

The contributions of this paper are three-fold. 
First, it introduces an end-to-end trainable pre-training network that allows to exploit language supervision to learn effective visual text representations. 
Second, we design a character-aware text encoder and a visual-textual decoder that can extract effective instance-level textual information and learn from partial text transcriptions without requiring text bounding boxes. 
Third, extensive experiments over multiple public datasets show that the proposed weakly supervised pre-trained network achieves superior performance on various scene text detection and spotting datasets.



\section{Related Work}

\subsection{Scene Text Detection and Spotting}
Most of recent scene text detectors are trained on fully-annotated data which can be broadly classified into two categories. The first category takes a bottom-up approach which first detects low-level text elements like characters \cite{baek2019character}, text segments \cite{Shi_2017_CVPR,tang2019seglink++} and text keypoints \cite{xue2022detection} and then groups them into words or text lines. The second category treats words as one specific type of objects, and many scene text detectors like EAST \cite{Zhou_2017_CVPR}, TextBoxes++ \cite{liao2018textboxes++}, RRD \cite{liao2018rotation} and PSENet \cite{wang2019shape} are designed to detect text bounding boxes directly with generic object detection or segmentation techniques. Besides, many researchers study the text-specific features for robust text detection through text border or counter \cite{xue2018accurate,wang2020contournet,zhu2021fourier,dai2021progressive}, deformation convolution \cite{wang2018geometry,xiao2020sequential},  local refinement \cite{zhang2019look,he2021most} and so on. Besides, many methods are designed to address the data bias. Some works \cite{gupta2016synthetic,zhan2018verisimilar,liao2020synthtext3d} aim to synthesize scene text images that can be used for training scene text detection, recognition and spotting models. In addition, WeText \cite{tian2017wetext} and OPM \cite{sun2019chinese} design different weakly supervised mechanisms to use different types of data for training. GA-DAN \cite{zhan2019ga} and TST \cite{wu2020synthetic} study the domain adaptation that adapt the synthetic scene text images to real. More recently, STKM \cite{wan2021self} is proposed to pre-train a general model backbone for different scene text detectors.

Besides, many end-to-end trainable scene text spotters have been designed in which text detector and recognizer are complementary to each other. Li et al. \cite{li2017towards} first integrates the scene text detector and RNN-based recognizer in to a unified network. Liu et al. \cite{liu2018fots} and He et al. \cite{he2018end} leverage more advanced scene text detectors or recognizers for better text spotting performances. More recently, Mask TextSpotters \cite{lyu2018mask,liao2019mask,liao2020mask} adopt Mask R-CNN \cite{he2017mask} as text detector and character segmentation or attention module for recognition. ABCNet \cite{liu2020abcnet,9525302} proposes to detect texts with Bezier curves. TextDragon \cite{feng2019textdragon} detects center lines of texts along which characters are recognized in sequence. Baek et al. \cite{baek2020character} proposes to detect characters by training with weakly supervised mechanism. Xing et al. \cite{xing2019convolutional} propose to detects and recognizes characters simultaneously. MANGO \cite{Qiao_Chen_Cheng_Xu_Niu_Pu_Wu_2021} is designed for text spotting with mask attention guidance. Additionally, text recognition with less annotation have been studied in \cite{tensmeyer2019training,baek2021if}.

\subsection{Vision-Language Pre-training}

As inspired by the advanced Transformer-based pre-training techniques \cite{devlin2018bert} in Natural Language Processing (NLP) community, many vision-language pre-training methods have been studied in recent years, which greatly promotes the many multi-modal tasks in computer vision community. ViLBERT \cite{lu2019vilbert} and LXMERT \cite{tan2019lxmert} present a two-stream framework with a vision-language co-attention module for cross-modal feature fusion. On the other hand, VisualBERT \cite{li2019visualbert}, Unicoder-VL \cite{li2020unicoder}, VL-BERT \cite{su2019vl}, and UNITER \cite{chen2019uniter} follow a single-stream framework (i.e. vanilla BERT structure), focusing on generic VL tasks including VCR and VQA. Besides, many VLP methods have been proposed for VL tasks such as RVL-BERT \cite{chiou2021visual} for visual relationship detection, PERVALENT \cite{hao2020towards} and VLN-BERT \cite{majumdar2020improving} for visual navigation, VisualID \cite{murahari2020large} and VD-BERT \cite{wang2020vd} for visual dialog, etc. 

\begin{figure}[t!]
  \centering
  \includegraphics[width=\linewidth]{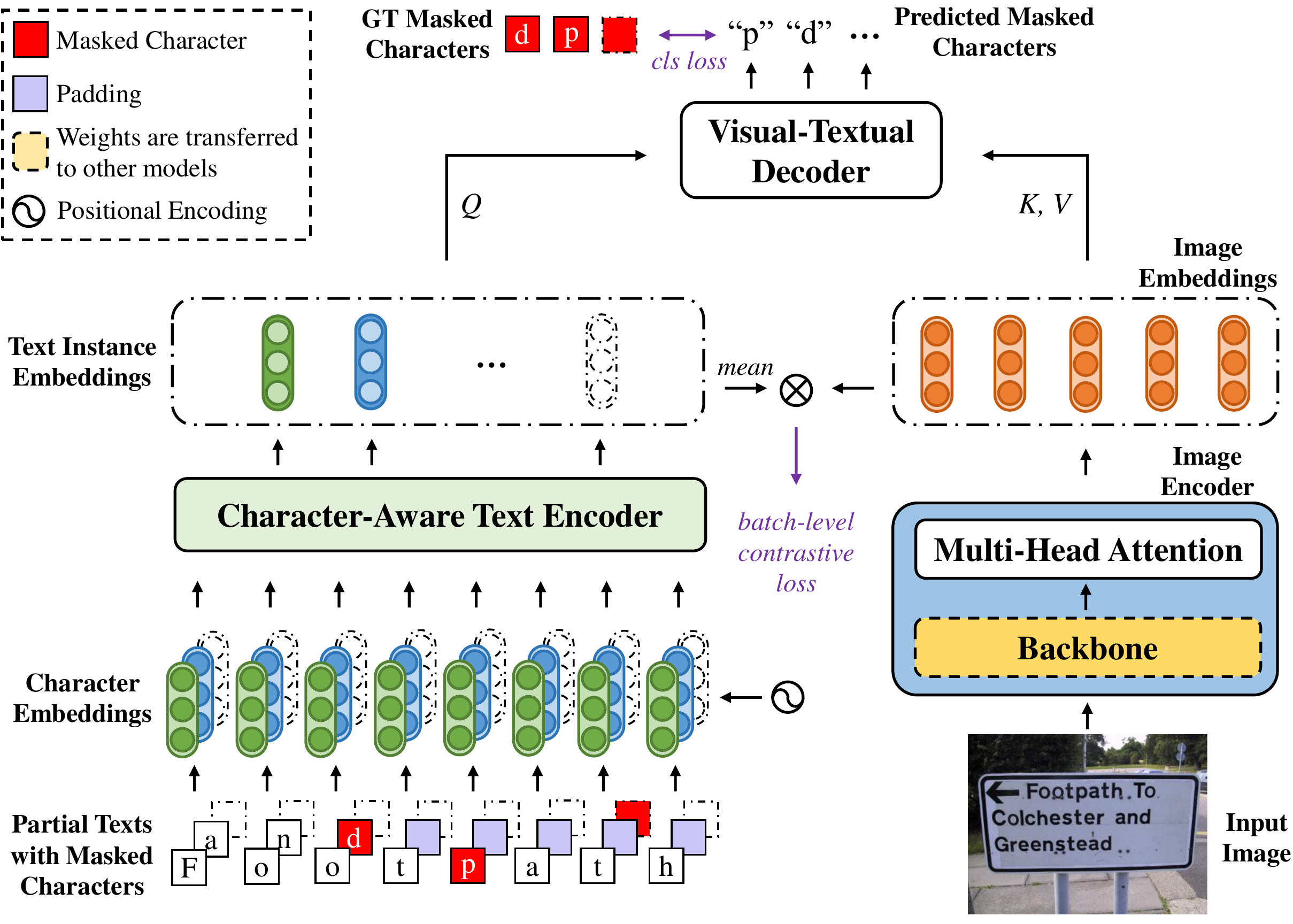}
  \caption{
  \textbf{The framework of the proposed method:} Given an input image, an image encoder (including a backbone followed by a multi-head attention layer) first extracts the visual features. Meanwhile, the characters in each text instance are transformed to character embeddings, and a character-aware text encoder further extracts text instance embeddings from the character embeddings. A visual-textual decoder models the interactions between the text instance embeddings and the corresponding image embeddings. During training, a random character in each text instance will be masked (as highlighted by red boxes) and the overall network is optimized by predicting the masked characters.
  }
  \label{fig:framework}
\end{figure}

\section{Methodology}



We present a pre-training technique that learns better scene text visual representations by feature alignment with textual information. As shown in Fig. \ref{fig:framework}, the proposed network first extracts image embeddings from input images by using an image encoder (including a network backbone ResNet-50 \cite{he2016deep} followed by a multi-head attention layer). A character-aware text encoder is designed to extract the textual information from the transcriptions of text instances in input images by encoding the sequence of characters in each text instance. The extracted textual and visual features are passed into a visual-textual decoder which models the interactions among the visual features of input image and the textual features of each individual text instance. During training, we randomly mask a character in each text instance and the network is optimized by predicting the masked characters leveraging the extracted visual and textual features.

\subsection{Character-Aware Text Encoder} \label{sec:text_encoder}

\begin{figure}[t!]
  \centering
  \includegraphics[width=\linewidth]{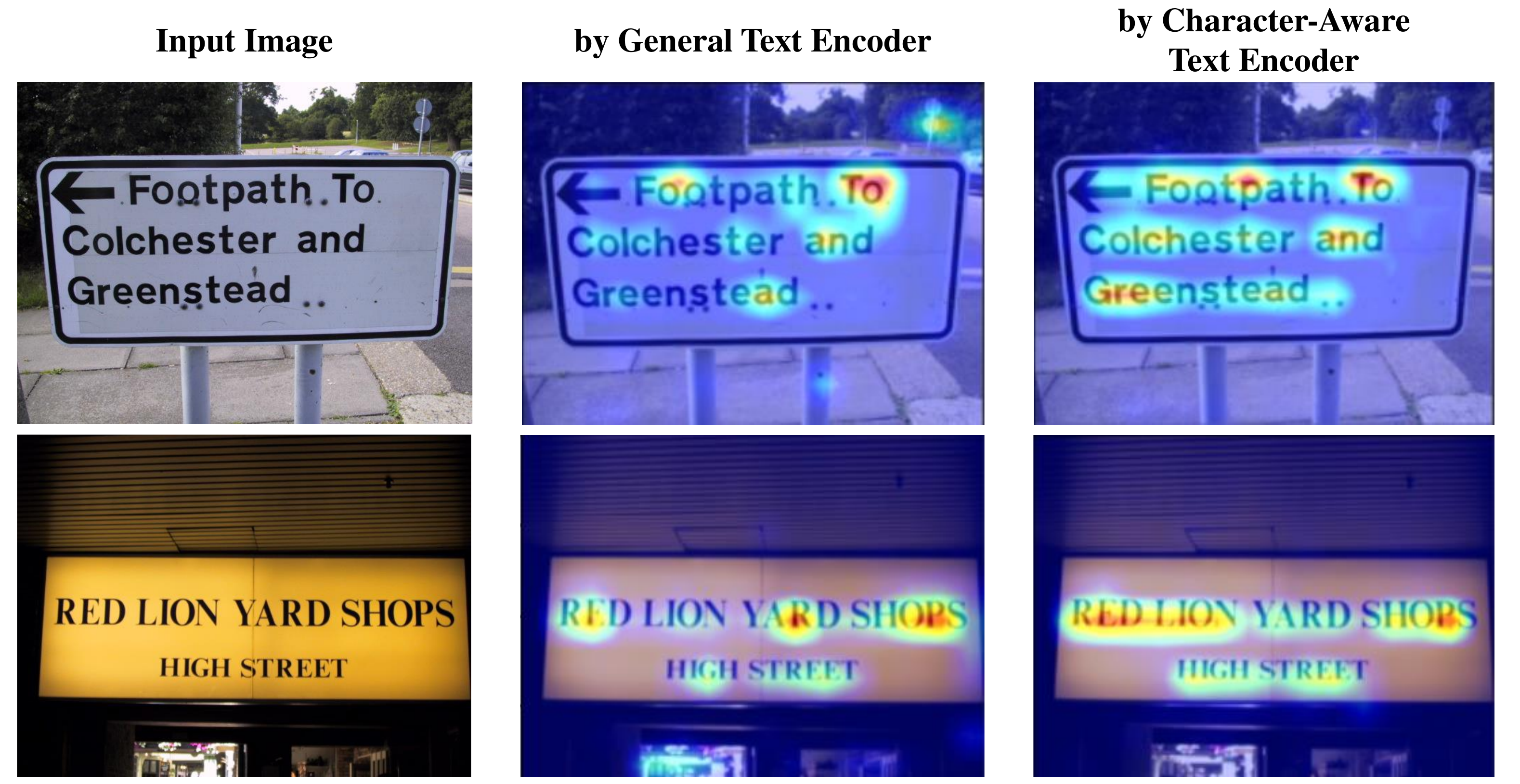}
  \caption{\textbf{Illustration of the proposed character-aware text encoder:} Given sample images in the first column, columns 2-3 show the attention maps (from the attention layer in the image encoder) that are obtained from models with the general sentence-level text encoder and the proposed character-aware text encoder, respectively. The proposed character-aware text encoder attends better to text regions as compared with the general text encoder, leading to better learning of the scene text visual representations of the network backbone.
  }
  \label{fig:encoder}
\end{figure}

In general VL tasks, texts (e.g. titles, captions, etc.) are usually sentences that consist of sequences of text tokens. As such, the text encoders for VL tasks are often designed to encode texts in a sequential way. However, the natural scene images in OCR tasks usually contain one or multiple text instances. The text tokens within each text instance are sequentially related to each other while those from different text instances are often completely irrelevant. This makes it difficult to encode these text instances by using a general text encoder. To address this issue, we design a character-aware text encoder.

The proposed character-aware text encoder extracts instance-level text embeddings with the input text instances as sequences of characters. Given $n$ annotated text instances $T = \{t_0, t_1, ... t_{n-1}\}$ in an image, each text instance $t_i$ consists of a sequence of characters $t_i = [c_0^i, c_1, ^i, ..., c_{k-1}^i]$. We embed the characters into fixed-sized vectors and add a set of learnt positional encoding \cite{vaswani2017attention} $PE = [PE_0, PE_1, ..., PE_k]$ to capture the sequential information of characters in each text instance only, which can be formulated by:
\begin{equation}
ce_j^i = W_c \cdot c_j^i + PE_j, \:\:\:\:\: i \in [0, n-1], \:\:\:\:\: j \in [0, k-1],
\end{equation}
where $W_c$ is the character embedding matrix. The encoded character embeddings of $i-th$ text instance $ce^i = [ce_0^i, ce_1^i, ..., ce_{k-1}^i]$ are hence passed into a Transformer \cite{vaswani2017attention} encoder which models the interaction among all characters in the text instance and extracts the text instance embeddings $te_i$ from its character embeddings $ce_i$. As a result, the character-aware text encoder extracts the text instance embeddings $te = \{te_0, te_1, ..., te_{n-1}\}$ from the annotated text instances $t = \{t_0, t_1, ... t_{n-1}\}$. Note a randomly selected character in each text instance is masked during training by setting it to the mask category.

The proposed character-aware text encoder effectively encodes the instance-level textual information and neglects the relations between each pair of text instances. In addition, it can help to learn better visual text representations. Fig. \ref{fig:encoder} shows two sample images accompanied with the attention maps from the attention layer in the image encoder (details in Fig. \ref{fig:framework}). As Fig. \ref{fig:encoder} shows, by extracting textual information from the general text encoder, the overall model only focuses on partial text instances (e.g. `Foo' and `th' of `Footpath'). This is because the tokens in general text encoder usually contain multiple characters (e.g. the token `Footpath' contains 8 characters) and the model thus tends to focus on the most important parts only in the token according to the linguistic knowledge. Instead, the proposed text encoder can attend better to all text regions in images with the awareness of each character, demonstrating the superiority of the proposed encoder on learning visual text representations for scene text detection and spotting tasks.

\subsection{Visual-Textual Decoder} \label{sec:mm_decoder}

The existing scene text pre-training techniques require fully-annotated data for training where the bounding boxes or transcriptions of all text instances are provided. However, such annotations are often extremely expensive and difficult to obtain. To address the data annotation bias, we present a visual-textual decoder that models the interaction between the input image and each individual annotated text while ignoring the unlabelled texts. The model thus can be trained by using the annotations of partial text instances in the images.

Given an input image $I$ as shown in Fig. \ref{fig:framework}, we first extract the image embeddings $ie$ and the the textual information $te$ by using an image encoder (including a network backbone followed by a multi-head attention layer) and a character-aware text encoder, respectively. The visual-textual decoder hence learns the relationships among $ie$ and each item in $te$ (i.e. embeddings of each text instance) to enhance the learning of visual representations. Specifically, the visual-textual decoder consists of 6 stacked decoder layers each of which contains a multi-head attention layer and a feed-forward network. The text instance embeddings $te$ are passed into the visual-textual decoder as queries and the image embeddings $ie$ are passed into the decoder as keys and values. This allows every text instance alone to attend over all positions in the image embeddings. Note that we don't adopt the self-attention layer in the visual-textual decoder in order to neglect the relationships between each pair of text instances and eliminates the effects of unlabelled text instances. The model thus can effectively learn from partial annotated text instances. Finally, the visual-textual decoder predicts the masked characters in each text instance for optimization. 

\begin{figure}[t!]
  \centering
  \includegraphics[width=\linewidth]{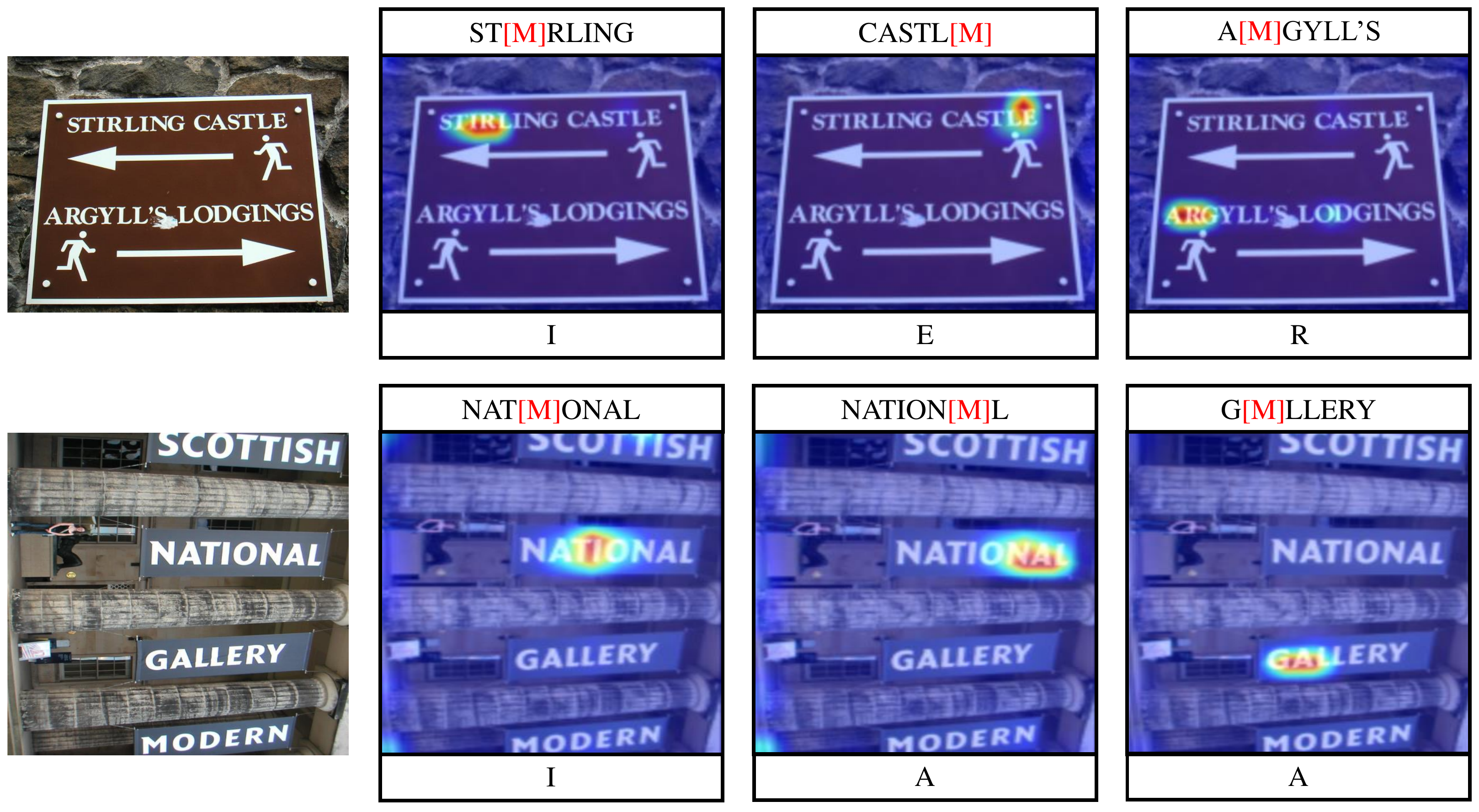}
  \caption{\textbf{Illustration of the proposed visual-textual decoder:} Given two sample images in the first column, the input text instances (masked characters are highlighted by \textcolor{red}{[M]}), corresponding attention maps in the decoder and the predicted masked characters are shown from top to bottom in each box in columns 2-4, respectively. The proposed visual-textual decoder aligns the visual and textual features well, which effectively attends and predicts the masked characters in images.
  }
  \label{fig:decoder}
\end{figure}

The masked characters can be predicted by learning the language knowledge from textual information only. We illustrate the attention maps of the decoder in Fig. \ref{fig:decoder} to demonstrate the effectiveness of the proposed visual-textual decoder. For each sample image in Fig. \ref{fig:decoder}, we pass three text instances (with masked characters \textcolor{red}{[M]}) into our network, and we obtain three attention maps and three predicted masked characters each of which corresponds to an input text instance. As Fig. \ref{fig:decoder} shows, the visual-textual decoder not only predicts the masked characters (e.g. `I' for `ST\textcolor{red}{[M]}RLING') but also attends the regions of corresponding masked characters well in the images. It can be seen that the proposed decoder aligns the visual and textual features to predict the masked characters (instead of using textual information alone), demonstrating the effectiveness of the proposed visual-textual decoder.

\subsection{Network Optimization} \label{sec:network_training}

During training, the proposed model takes text instances $T$ (with masked characters $\bm{y}^{msk}$) and images $I$ as inputs, and predicts the masked characters $\bm{p}^{msk}(I, T)$ for optimization. We consider the masked character prediction as a classification problem and adopt cross-entropy H for optimization:

\begin{equation}
\mathcal{L}_{cls} =  \mathbb{E}_{(I, T)\sim D}\text{H}(\bm{y}^{msk}, \bm{p}^{msk}(I, T)).
\end{equation}

Besides, as inspired by CLIP \cite{radford2021learning}, we adopt a batch-level contrastive loss for faster convergence. Given $N$ images and $N$ texts in a training batch, we form $N^2$ (text, image) pairs from all texts and images, where $N$ pairs of texts and images are correlated with each other and $N^2 - N$ pairs are unrelated. For each image and text, we calculate the softmax-normalized image-to-text and text-to-image
similarity as:
\begin{equation}
p^{i2t}_b(I) = \frac{\text{exp}(I, T_b)}{ {\textstyle \sum_{b=1}^{B}} \text{exp}(I, T_b)}, \:\:\:\:\:p^{t2i}_b(T) = \frac{\text{exp}(T, I_b)}{ {\textstyle \sum_{b=1}^{B}} \text{exp}(T, I_b)}.
\end{equation}
Let $\bm{y}^{i2t}(I)$ and $\bm{y}^{t2i}(T)$ denote the ground-truth one-hot similarity, where negative pairs have a probability of 0 and the positive pair has a probability of 1. The batch-level contrastive loss is thus defined by:

\begin{equation}
\mathcal{L}_{bc} =  \mathbb{E}_{(I, T)\sim D}[\text{H}(\bm{y}^{i2t}(I), \bm{p}^{i2t}(I)) + \text{H}(\bm{y}^{t2i}(T), \bm{p}^{t2i}(T))].
\end{equation}

The full pre-training objective is defined by:

\begin{equation}
\mathcal{L}=\mathcal{L}_{cls}+\mathcal{L}_{bc}.
\end{equation}

\section{Experiments}

\subsection{Datasets}
We use a number of public datasets in our experiments including SynthText\cite{gupta2016synthetic}, ICDAR2019-LSVT\cite{sun2019icdar}, CTW1500\cite{yuliang2017detecting}, Total-Text\cite{ch2017total}, and ICDAR2015\cite{karatzas2015icdar}. More details are available in the supplementary material.






\subsection{Implementation Details}
\noindent\textbf{Pre-training:} We use ResNet-50 \cite{he2016deep} as the backbone in the image encoder of the proposed network. The input images are resized to $512 \times 512$ during training. We adopt the Adam optimizer \cite{kingma2014adam} with decoupled weight decay regularization \cite{loshchilov2017decoupled} applied to all weights that are not gains or biases. The initial learning rate is $1e^{-4}$ which decays using a cosine schedule \cite{loshchilov2016sgdr}. The model is trained end-to-end for 100 epochs on 8 Tesla V100 GPUs with batch size of 640. The length of each text instance is set as 25 following \cite{yu2020towards,xue2021i2c2w}. 

\noindent\textbf{Fine-tuning:} We fine-tune several scene text detectors and spotters for evaluation of the proposed method including: 1) PSENet \cite{wang2019shape}, 2) DB \cite{liao2020real}, 3) FCENet \cite{zhu2021fourier}, 4) TextBPN \cite{zhang2021adaptive}, and 5) Mask TextSpotter-v3 \cite{liao2020mask}. 
More details are available in the supplementary material.


\subsection{Experimental Results}

We evaluate the proposed method from three aspects. First, we evaluate the performances of the proposed method by training with weakly annotated data (i.e. with partial annotated text instances available in each image). Second, we compare the proposed method with the existing pre-training techniques in scene text community. Third, we compare the proposed method with the state-of-the-art scene text detectors and spotters.

\addtolength{\tabcolsep}{5pt} 
\begin{table}[!t]
\centering
\caption{\textbf{Scene text detection} performances of different models on \textbf{ICDAR2019-LSVT} dataset. `+oCLIP': Our pre-trained model with 400,000 weakly annotated images in ICDAR2019-LSVT dataset is adopted for fine-tuning.}
\begin{tabular}{lccc}
\toprule
\textbf{Model}                          &\textbf{Precision} &\textbf{Recall}      &\textbf{F-score} \\
\midrule
MSR  \cite{xue2019msr}                  &  86.4             &   63.4              &   73.1               \\
Keypoint  \cite{xue2022detection}       &  78.5             &   70.1              &   74.1           \\

\midrule
DB \cite{liao2020real}             &  76.5            &   70.5               &    73.4              \\
\textbf{DB+oCLIP}  &  \textbf{81.5}    &   \textbf{70.9}         &    \textbf{75.8}             \\
\midrule
PSENet  \cite{wang2019shape}            &  90.4             &   63.5          &    74.6          \\
\textbf{PSENet+oCLIP}                            &  \textbf{90.7}      &   \textbf{67.0}        &    \textbf{77.1}        \\
\bottomrule
\end{tabular}
\label{tab:lsvt_det}
\end{table}
\addtolength{\tabcolsep}{-5pt} 

\addtolength{\tabcolsep}{2pt} 
\begin{table}[!t]
\centering
\caption{\textbf{Scene text spotting} performances of different models on \textbf{ICDAR2019-LSVT} dataset. `+oCLIP': Our pre-trained model with 400,000 weakly annotated images in ICDAR2019-LSVT dataset is adopted for fine-tuning. `P', `R', `F', `1-NED', and `E2E' refer to Precision, Recall, F-score, Normalized metric in terms of Normalized Edit Distance, and end-to-end, respectively.}
\begin{tabular}{lccccccc}
\toprule
\multirow{2}{*}{\textbf{Method}}  & \multicolumn{3}{c}{\textbf{Detection}} & \multicolumn{4}{c}{\textbf{E2E Spotting}}             \\ \cline{2-8} 
                                  & \textbf{P}  & \textbf{R}  & \textbf{F} & \textbf{1-NED} & \textbf{P} & \textbf{R} & \textbf{F} \\
\midrule
Mask TextSpotter-V3               &  80.5           &  61.0         &  69.4      &      35.7          &  32.1          &     24.4       &     27.7       \\
\textbf{Mask TextSpotter-V3+oCLIP} &  \textbf{80.6}     & \textbf{61.9}      &  \textbf{70.1}     &  \textbf{39.0}        &  \textbf{37.4}     &   \textbf{28.7}    & \textbf{32.5}    \\
\bottomrule
\end{tabular}
\label{tab:lsvt_spot}
\end{table}
\addtolength{\tabcolsep}{-2pt}

\subsubsection{Weakly Supervised Pre-training:} We evaluate the performances of the proposed method on learning visual text representations from weakly annotated data. We first conduct the experiments by pre-training our model on 400,000 weakly annotated images (i.e. only the transcription of the text-of-interest in each image is provided), and fine-tuning different scene text detectors and spotters on 30,000 fully annotated images from ICDAR2019-LSVT dataset. As Table \ref{tab:lsvt_det} and \ref{tab:lsvt_spot} show, the proposed method improves the performances of different scene text detectors and spotters, demonstrating that the proposed method effectively learns the visual representations from weakly annotated data. Note that most previous approaches are designed to train on fully annotated images and they can't utilize the weakly annotated images from ICDAR2019-LSVT dataset well.

\addtolength{\tabcolsep}{5pt} 
\begin{table}[!t]
\centering
\caption{The effectiveness of the proposed weakly supervised pre-training technique: We pre-train four models by using different proportions of text instances in SynthText dataset (e.g. 1 out of 4 text instances in each image are used for training for `25\%' model), and transfer the models weights to fine-tune PSENet on Total-Text dataset. `Baseline': Train PSENet on SynthText and then fine-tune on Total-Text.}
\begin{tabular}{cccc}
\toprule
\textbf{\% annotated texts} & \textbf{Precision} & \textbf{Recall} & \textbf{F-score} \\
\midrule
No Pre-train                        &  81.8    &  75.1    & 78.3           \\
Baseline                        &  87.8    &  79.0   & 82.6            \\
\midrule
25\%                         &  90.2               &  80.1          &  84.8                \\
50\%                         &   91.1            &  80.0        &   85.2            \\
75\%                         &   90.6                 &  80.8         &     85.4             \\
\textbf{100\%}                        &   \textbf{90.7}      &    \textbf{80.8}      &    \textbf{85.5}      \\
\bottomrule
\end{tabular}
\label{tab:ratio_weak_annotation}
\end{table}
\addtolength{\tabcolsep}{-5pt} 

In addition, we conduct an experiment on SynthText dataset to show the effects of the amount of annotated texts on model performances. We first prepare four sets of text annotations from SynthText dataset by randomly selecting different proportions of text instances (i.e. 25\%, 50\%, 75\%, and 100\%) in each image (e.g. 1 out of 4 text instances in each image are used for training `25\%' model). Next, we pre-train four models on all images in SynthText dataset by using the four sets of text annotations, and then transfer the backbone weights to fine-tune PSENet on Total-Text dataset. For comparison, we report the performances of two additionally models including: 1) `No Pre-train' model in which no pre-training is adopted, and 2) `Baseline' model that first trains PSENet on SynthText and then fine-tunes on Total-Text, respectively. As Table \ref{tab:ratio_weak_annotation} shows, all four pre-train models help to improve the performances of PSENet, which outperforms the `No Pre-train' and `Baseline' models significantly. Besides, the four models achieve comparable performances on scene text detection task by pre-training on different amount of annotated texts, demonstrating the effectiveness of the proposed weakly supervised learning.

\addtolength{\tabcolsep}{3pt} 
\begin{table}[!t]
\centering
\caption{Comparison with existing scene text pre-training techniques: by pre-training on the same set of data (i.e. SynthText dataset), the proposed pre-training method outperforms the existing pre-training techniques consistently across different datasets. `+SynthText': Train PSENet with SynthText and then fine-tune with Total-Text.}
\begin{tabular}{lcccccc}
\toprule
\multirow{2}{*}{Model} & \multicolumn{3}{c}{\textbf{Total-Text}} & \multicolumn{3}{c}{\textbf{CTW1500}} \\\cline{2-7} 
                       & \textbf{P}        & \textbf{R}        & \textbf{F}        & \textbf{P}       & \textbf{R}       & \textbf{F}       \\
\midrule
PSENet \cite{wang2019shape}  & 81.8    &  75.1    & 78.3     &  80.6     &  75.6            & 78.0    \\
PSENet+SynthText       &  87.8    &  79.0   & 82.6    &  81.8    &  77.8    & 79.7       \\
PSENet+STKM\cite{wan2021self}            &  86.3    &  78.4   & 82.2    & 85.1     &  78.2            & 81.5          \\
PSENet+oCLIP[SynthText]  &   90.7       & 80.8       & 85.5     &  86.3       & 79.6    &    82.8     \\
\textbf{PSENet+oCLIP[Web Images]}  & \textbf{92.2}    & \textbf{82.4}   &  \textbf{87.0}    &  \textbf{87.5}       &  \textbf{79.9}   &  \textbf{83.5}       \\
\bottomrule
\end{tabular}
\label{tab:compare_pretrain}
\end{table}
\addtolength{\tabcolsep}{-1pt}

\subsubsection{Comparing with Existing Scene Text Pre-training Strageties:} We compare the proposed method with two scene text pre-training strategies including: (1) training PSENet on SynthText dataset and then fine-tuning on real dataset, and (2) pre-training on SynthText by using STKM \cite{wan2021self} and transferring the pre-trained weights to fine-tune PSENet on real dataset. For a fair comparison, we pre-train our model on SynthText with full annotations and transfer the backbone weights for fine-tuning PSENet on real datasets. As Table \ref{tab:compare_pretrain} shows, by pre-training on the same set of data, the proposed method outperforms the existing pre-training techniques by +3.3\% and +1.3\% in F-score on Total-Text and CTW1500 datasets, respectively.

\subsubsection{Automatic Data Acquisition and Training from Web Images:} 
The proposed oCLIP can be simply applied to an automatic data acquisition and training pipeline due to the success of learning from weakly-annotated images. We extracted texts from 40 million web images and filtered out less-confident ones by using the existing scene text detector and recognizer form model pre-training. As Table \ref{tab:compare_pretrain} shows, by learning from the automatically extracted data from web images, oCLIP significantly improves the performances of PSENet on Total-Text and CTW1500 datasets. More details are available in supplementary material.

\subsubsection{Comparing with State-of-the-Art Scene Text Detectors and Spotters:} 

\addtolength{\tabcolsep}{2pt} 
\begin{table}[!t]
\centering
\caption{Comparison with state-of-the-art scene text detection techniques on \textbf{CTW1500} dataset. `+oCLIP' refers to that our pre-trained model on SynthText dataset is adopted for fine-tuning. `RN50', `PD', `Syn', and `MLT' refer to ResNet-50, pre-training data, SynthText dataset, and ICDAR2027-MLT dataset, respectively}
\begin{tabular}{lcccl}
\toprule
\textbf{Model}        &\textbf{PD}    &\textbf{Precision}    &\textbf{Recall}   &\textbf{F-score} \\
\midrule
TextSnake \cite{long2018textsnake}      & Syn          &  67.9                &  \textbf{85.3}            & \ \ \;75.6           \\
ATRR \cite{wang2019arbitrary}           & -          &  80.1                &  80.2            & \ \ \;80.1           \\
TextField  \cite{xu2019textfield}      & Syn          &  83.0                &  79.8            & \ \ \;81.4          \\
Keypoint  \cite{xue2022detection}       & Syn          &  \textbf{88.3}                &  77.7            & \ \ \;82.7           \\
PAN  \cite{wang2019efficient}          & Syn          &  88.0                &  79.4            & \ \ \;83.5           \\
CRAFT  \cite{baek2019character}         & Syn          &  86.4                &  81.4            & \ \ \;83.7           \\
ContourNet  \cite{wang2020contournet}   & -          &  83.7                &  84.1            & \ \ \;83.9           \\
SD \cite{xiao2020sequential}            & MLT        &  85.8                &  82.3            & \ \ \;84.0              \\
DRRG \cite{zhang2020deep}           & MLT        &  85.9                &  83.0            & \ \ \;84.5              \\
TextBPN \cite{zhang2021adaptive}           & Syn        &  87.8                &  81.5            & \ \ \;84.5              \\
\midrule
DB-RN50 \cite{liao2020real}           & -          &  81.1                &  80.6           & \ \ \;80.8              \\ 
\textbf{DB-RN50+oCLIP}                 & Syn          &  82.5       & 81.5    & \ \ \;\textbf{82.0 \textcolor{blue}{(+1.2)}}             \\\midrule
FCENet-RN50  \cite{zhu2021fourier}           & -          &  85.7                &  80.7            & \ \ \;83.1           \\
\textbf{FCENet-RN50+oCLIP}      & Syn         &  87.2                &  83.9            & \ \ \;\textbf{85.6 \textcolor{blue}{(+2.5)}}           \\
\bottomrule
\end{tabular}
\label{tab:ctw_det}
\end{table}
\addtolength{\tabcolsep}{-2pt}

\addtolength{\tabcolsep}{2pt} 
\begin{table}[!t]
\centering
\caption{Comparison with state-of-the-art scene text detection techniques on \textbf{Total-Text} dataset. `+oCLIP' refers to that our pre-trained model on SynthText dataset is adopted for fine-tuning. `RN50', `PD', `Syn', and `MLT' refer to ResNet-50, pre-training data, SynthText dataset, and ICDAR2027-MLT dataset, respectively}
\begin{tabular}{lcccl}
\toprule
\textbf{Model}                 &\textbf{PD}    &\textbf{Precision}    &\textbf{Recall}   &\textbf{F-score} \\
\midrule
TextSnake \cite{long2018textsnake}    & Syn          &  82.7                &  74.5            & \ \ \;78.4           \\
ATRR \cite{wang2019arbitrary}        & -          &  80.9                &  76.2            & \ \ \;78.5           \\
MSR  \cite{xue2019msr}               & Syn          &  83.8                &  74.8            & \ \ \;79.0           \\
TextField  \cite{xu2019textfield}   & Syn          &  81.2                &  79.9            & \ \ \;80.6           \\
PAN  \cite{wang2019efficient}         & Syn          &  88.0                &  79.4            & \ \ \;83.5           \\
CRAFT  \cite{baek2019character}       & MLT          &  87.6                &  79.9            & \ \ \;83.6           \\
Keypoint  \cite{xue2022detection}     & Syn          &  86.1                &  82.6            & \ \ \;84.4           \\
ContourNet  \cite{wang2020contournet}  & -          &  86.5                &  84.9            & \ \ \;85.4           \\
DRRG \cite{zhang2020deep}          & MLT        &  86.5                &  84.9            & \ \ \;85.7              \\
SD \cite{xiao2020sequential}           & MLT        &  \textbf{89.2}                &  84.7            & \ \ \;86.9              \\
\midrule
DB-RN50 \cite{liao2020real}         & -          &  81.7                &  75.0            & \ \ \;78.2              \\
\textbf{DB-RN50+oCLIP}                & Syn          &  86.1       & 82.1    & \ \ \;\textbf{84.1 \textcolor{blue}{(+5.9)}}   \\  
\midrule
TextBPN \cite{zhang2021adaptive}          & -        &  88.0                &  82.9            & \ \ \;85.4              \\
\textbf{TextBPN+oCLIP}         & Syn          & 89.0           &  \textbf{85.3}        &  \ \ \;\textbf{87.1 \textcolor{blue}{(+1.7)}}          \\
\bottomrule
\end{tabular}
\label{tab:total_det}
\end{table}
\addtolength{\tabcolsep}{-2pt}

\addtolength{\tabcolsep}{2pt} 
\begin{table}[!t]
\centering
\caption{Comparison with state-of-the-art scene text detection techniques on \textbf{ICDAR2015} dataset. `+oCLIP' refers to that our pre-trained model on SynthText dataset is adopted for fine-tuning. `RN50', `PD', `Syn', and `MLT' refer to ResNet-50, pre-training data, SynthText dataset, and ICDAR2027-MLT dataset, respectively.}
\begin{tabular}{lcccl}
\toprule
\textbf{Model}                     &\textbf{PD}    &\textbf{Precision}    &\textbf{Recall}   &\textbf{F-score} \\
\midrule
SegLink \cite{shi2017detecting}    & Syn          &  76.1                &  76.8            & \ \ \;75.0           \\
TextField  \cite{xu2019textfield}     & Syn          &  84.3                &  80.1            & \ \ \;82.4          \\
TextSnake \cite{long2018textsnake}     & Syn          &  84.9                &  80.4            & \ \ \;82.6           \\
PAN  \cite{wang2019efficient}       & Syn          &  84.0                &  81.9            & \ \ \;82.9           \\
ATRR \cite{wang2019arbitrary}         & -          &  90.4                &  83.3            & \ \ \;86.8           \\
CRAFT  \cite{baek2019character}     & MLT          &  89.8               &  84.3            & \ \ \;86.9           \\
ContourNet  \cite{wang2020contournet}   & -          &  87.6                &  86.1            & \ \ \;86.9           \\
SD \cite{xiao2020sequential}         & MLT        &  88.7                &  \textbf{88.4}            & \ \ \;88.6              \\
\midrule
DB-RN50 \cite{liao2020real}        & -          &  89.3                &  74.0            & \ \ \;80.9              \\
\textbf{DB-RN50+oCLIP}               & Syn          &  89.1       & 82.0    & \ \ \;\textbf{85.4 \textcolor{blue}{(+4.5)}}             \\
\midrule
FCENet-RN50  \cite{zhu2021fourier}        & -          &  88.0                &  81.9            & \ \ \;84.9           \\
\textbf{FCENet-RN50+oCLIP}       & Syn          &  \textbf{91.2}                &  82.7            & \ \ \;\textbf{86.7 \textcolor{blue}{(+1.8)}}           \\
\bottomrule
\end{tabular}
\label{tab:ic15_det}
\end{table}
\addtolength{\tabcolsep}{-2pt} 

\addtolength{\tabcolsep}{5pt} 
\begin{table}[!t]
\centering
\caption{Comparison with state-of-the-art scene text spotting techniques on \textbf{ICDAR2015} and \textbf{Total-Text} dataset. `+oCLIP' refers to that the model are fine-tuned from the our pre-trained model on SynthText dataset. `S', `W', and `G' refer to end-to-end recognition with strong, weak, generic lexicon for ICDAR2015. `Full' refers to full lexicon for Total-Text.
}
\begin{tabular}{lcccc}
\toprule
\textbf{Model}                    & \multicolumn{3}{c}{\textbf{ICDAR2015}}        & \textbf{Total-Text} \\ \cline{2-5} 
\textbf{}                         & \textbf{S}    & \textbf{W}    & \textbf{G}    & \textbf{Full}       \\
\midrule
CharNet \cite{xing2019convolutional}                & 80.1          & 74.5          & 62.2          & -                   \\
FOTS  \cite{liu2018fots}                            & 83.6          & 74.5          & 62.2          & -                   \\
TextDragon   \cite{feng2019textdragon}              & 82.5          & 78.3          & 65.2          & 74.8                \\
Boundary TextSpotter   \cite{wang2020all}           & 79.7          & 75.2          & 64.1          & -                   \\
PAN++   \cite{wang2021pan++}                        & 82.7          & 78.2          & 69.2          & 78.6                \\
ABCNet-V2  \cite{9525302}                           & 82.7          & 78.5          & 73.0          & 78.1                \\
\midrule
Mask TextSpotter-V3  \cite{liao2020mask}            & 83.3          & 78.1          & 74.2          & 78.4                \\
\textbf{Mask TextSpotter-V3+oCLIP}                   & \textbf{84.1} & \textbf{78.6} & \textbf{74.3} & \textbf{79.6}     \\
\bottomrule
\end{tabular}
\label{tab:spot}
\end{table}
\addtolength{\tabcolsep}{-5pt}

We further conduct experiments to compare the proposed method with state-of-the-art scene text detection and spotting techniques. For a fair comparison, we pre-train a model by our method on SynthText with full annotations and transfer the backbone weights to fine-tune DB, FCENet, TextBPN, and Mask TextSpotter-V3 on real datasets. As Table \ref{tab:ctw_det}-\ref{tab:spot} show, the proposed pre-trained model effectively promote the existing scene text detectors to state-of-the-art performances on different dataset. In addition, by transferring the pre-trained weights from our model, the performances of different scene text detectors and spotters are consistently improved by large margins.

\subsection{Ablation Studies}

\addtolength{\tabcolsep}{4pt} 
\begin{table}[!t]
\centering
\caption{\textbf{Ablation study} of the proposed method for scene text detection over Total-Text dataset. We fine-tune PSENet by using the pre-trained models with different modules. `CAE', `VTD', and `BCL' refer to character-aware encoder, visual-textual decoder, and batch-level contrastive loss, respectively. }
\begin{tabular}{ccccccc}
\toprule
\textbf{}   & \textbf{CAE} & \textbf{VTD} & \textbf{BCL} & \textbf{Precision} & \textbf{Recall} & \textbf{F-score} \\
\midrule
No Pretrain &              &              &             & 81.8       & 75.1       & 78.3       \\
1           &              &              & \checkmark  & 88.1       & 77.7       & 82.6       \\
2           & \checkmark   &              & \checkmark  & 89.6       & 78.9       & 83.9       \\
3           & \checkmark   & \checkmark   &             & 89.3       & 77.4       & 82.9       \\
4           & \checkmark   & \checkmark   & \checkmark  & \textbf{90.7}       & \textbf{80.8}       & \textbf{85.5}     \\
\bottomrule
\end{tabular}
\label{tab:ablation}
\end{table}
\addtolength{\tabcolsep}{-4pt}

We study the contributions of different modules in our method including a character-aware encoder (CAE), a visual-textual decoder (VTD), and a batch-level contrastive loss (BCL). We train four models with different modules included on fully annotated SynthText dataset and fine-tune PSENet on Total-Text dataset. As Table \ref{tab:ablation} shows, with the inclusion of different modules in our network, the performances of PSENet can be improved consistently, demonstrating the effectiveness of different modules in of network.

\section{Conclusion}
This paper presents a weakly supervised pre-training technique for scene text detection and spotting tasks. It focuses on the joint learning of visual and textual information from images and text transcriptions to enhance the learning of visual representations. It designs a character-aware text encoder and a visual-textual decoder that improves the feasibility of the proposed method on learning from partial text transcriptions only without text bounding boxes. Experimental results show that the proposed method can effectively learn from weakly-annotated scene text datasets which greatly mitigates the data acquisition challenge and significantly promotes different scene text detectors and spotters.

%
%
\bibliographystyle{splncs04}
\bibliography{egbib}
\end{document}


\pagestyle{headings}
\mainmatter
\def\ECCVSubNumber{1958}  


\title{Language Matters: A Weakly Supervised Vision-Language Pre-training Approach for Scene Text Detection and Spotting}

\author{Chuhui Xue\and
Wenqing Zhang  \and
Yu Hao \and \\
Shijian Lu  \and
Philip Torr \and
Song Bai }
\authorrunning{Xue et al.}
\institute{Supplementary Material}

\maketitle

\section{Automatic Data Acquisition and Training from Web Images}

Most existing scene text detection and spotting models are trained on fully-annotated data that are difficult to obtain from web images. Instead, the proposed weakly supervised pre-training approach can be simply applied to an automatic data acquisition and training pipeline by: (1) Extracting texts from web images by the existing OCR techniques; (2) Filtering out the less confident text instances (i.e. detected and recognized texts with low confident scores); (3) Pre-training a model on the collected web images and extracted text instances.

\addtolength{\tabcolsep}{3pt}
\begin{table}[!h]
\centering
\caption{Automatic data acquisition and training from web images: By pre-training on the automatically extracted images and texts from web, the proposed method can promote the existing scene text detectors significantly on TotalText and CTW1500 datasets.}
\begin{tabular}{lcccccc}
\toprule
\multirow{2}{*}{Model} & \multicolumn{3}{c}{\textbf{Total-Text}} & \multicolumn{3}{c}{\textbf{CTW1500}} \\\cline{2-7} 
                       & \textbf{P}        & \textbf{R}        & \textbf{F}        & \textbf{P}       & \textbf{R}       & \textbf{F}       \\
\midrule
PSENet \cite{wang2019shape}  & 81.8    &  75.1    & 78.3     &  80.6     &  75.6            & 78.0    \\
PSENet+SynthText       &  87.8    &  79.0   & 82.6    &  81.8    &  77.8    & 79.7       \\
PSENet+Ours[SynthText]  &   90.7       & 80.8       & 85.5     &  86.3       & 79.6    &    82.8     \\
\textbf{PSENet+Ours[Web Images]}  & \textbf{92.2}    & \textbf{82.4}   &  \textbf{87.0}    &  \textbf{87.5}       &  \textbf{79.9}   &  \textbf{83.5}       \\
\bottomrule
\end{tabular}
\label{tab:web_image}
\end{table}
\addtolength{\tabcolsep}{-3pt}

We conduct an experiment following this pipeline. We first extract texts from web images by using PSENet \cite{wang2019shape} for detection and Conformer \cite{gulati2020conformer} for recognition. Then, we filter out the less confident texts and non-text images, resulting in 40 million image-text pairs. Finally, we pre-train a model by using the proposed method and transfer the weights in the pre-trained model to fine-tune PSENet on Total-Text and CTW1500 datasets. As Table \ref{tab:web_image} shows, by automatically extracting data and pre-training, the proposed method significantly improves the performances of PSENet on Total-Text and CTW1500 datasets, demonstrating the effectiveness of the proposed method. This result also shows that the scene text models can be effectively promoted by large-scale pre-training on web images.

\begin{figure}[t!]
  \centering
  \includegraphics[width=\linewidth]{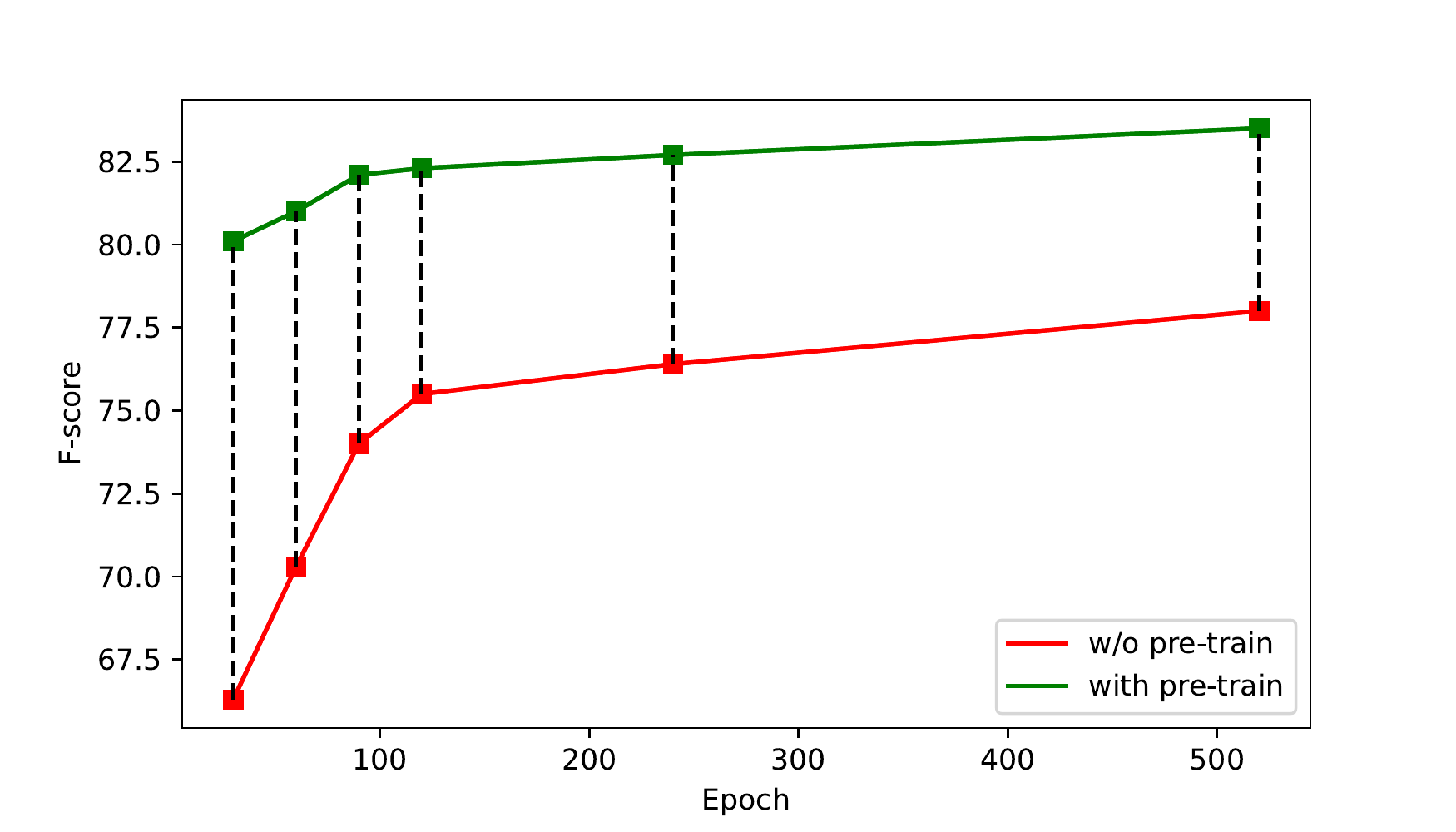}
  \caption{By pre-training on web images, the model converges faster than the original model without pre-training.
  } 
  \label{fig:convergence}
\end{figure}

Besides, the proposed pre-trained models effectively accelerate the convergence of the scene text model. As Fig. \ref{fig:convergence} shows, the scene text detector with pre-trained converges faster than the original model without pre-training.

\section{Datsets}
\noindent\textbf{SynthText} \cite{gupta2016synthetic} contains more than 800,000 synthetic scene text images most of which are at word level with multi-oriented rectangular annotations. The texts are in English in SynthText dataset.

\noindent\textbf{ICDAR2019-LSVT} \cite{sun2019icdar} consists of 450,000 images with mostly Chinese texts. 400,000 images are weakly annotated in which only the transcription of the text-of-interest in these images is provided. Besides, 50,000 images are fully annotated which are split into a training set with 30,000 images and a test set with 20,000 images.

\noindent\textbf{CTW1500} \cite{yuliang2017detecting} consists of 1,000 training images and 500 test images that contain 10,751 multi-oriented text instances of which 3,530 are arbitrarily curved. Most of the text instances are annotated at text-line level by using 14 vertices, where texts are largely in English and Chinese.

\noindent\textbf{Total-Text} \cite{ch2017total} consists of 1,255 training images and 300 test images where texts are all in English. It contains a large number of multi-oriented curved text instances each of which is annotated at word level by using a polygon. 

\noindent\textbf{ICDAR2015} \cite{karatzas2015icdar} has 1000 training images and 500 test images which are collected by Google Glass and suffers from low resolution and motion blur. All text instances are annotated at word level using quadrilateral boxes.

\begin{figure}[t!]
  \centering
  \includegraphics[width=\linewidth]{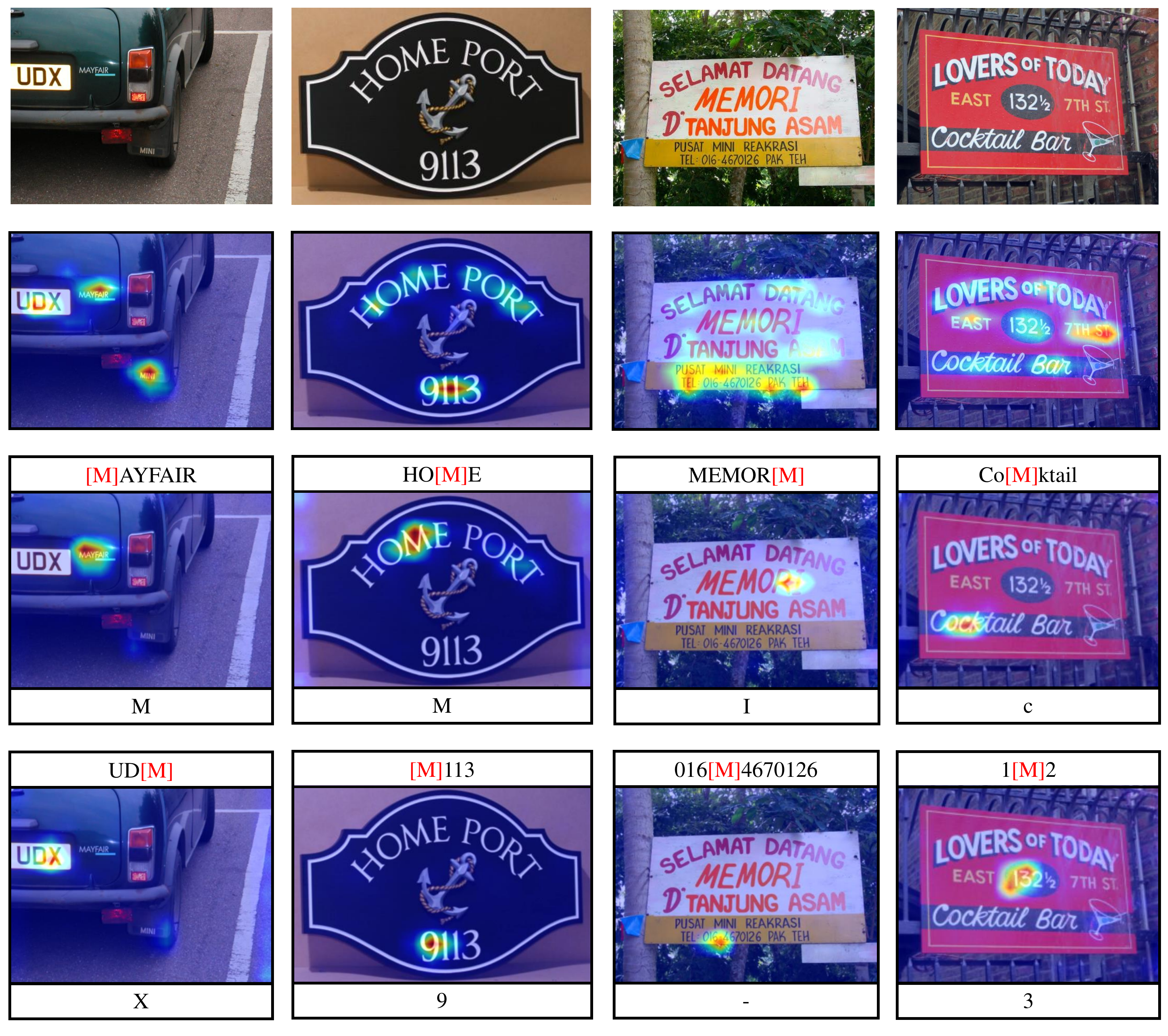}
  \caption{Given sample images in the first row, the second row shows the corresponding attention maps in the image encoder. Rows 3-4 shows a context text and a contextless text as input, respectively, as well as the corresponding attention maps in the decoder and the predicted characters. The encoder and decoder effectively attend to the text and character regions, respectively.
  } 
  \label{fig:more}
\end{figure}
\section{Implementation Details}
We fine-tune several scene text detectors and spotters for evaluation of the proposed method including: 1) PSENet \footnote{https://github.com/whai362/PSENet} \cite{wang2019shape}, 2) DB \footnote{https://github.com/MhLiao/DB} \cite{liao2020real}, 3) FCENet \footnote{https://github.com/open-mmlab/mmocr} \cite{zhu2021fourier}, 4) TextBPN \footnote{https://github.com/GXYM/TextBPN} \cite{zhang2020deep}, and 5) Mask TextSpotter-v3 \footnote{https://github.com/MhLiao/MaskTextSpotterV3} \cite{liao2020mask}. The experiments are conducted by using the corresponding open-source codes. For DB, we replace the original network backbone (i.e. deformable ResNet-50) with ResNet-50 for better demonstration of the proposed method. For TextBPN, we follow the experimental settings reported in their paper to re-train the overall the model as the configuration files are not provided. 

\section{More Samples}

The proposed method can attend to text regions with character awareness with language supervision only. Fig. \ref{fig:more} shows four more sample images as well as their attention maps. As Fig. \ref{fig:more} shows, the proposed model can effectively attend to the text regions and the missing character regions (corresponding to each input text instance). Especially in the last row of Fig. \ref{fig:more}, the proposed decoder can attend to the regions of missing characters from contextless text instance, demonstrating the effectiveness of the proposed method on modelling the relations of visual and textual information.

\section{More Experimental Results}
Recent scene text spotters are usually evaluated on ICDAR2015 dataset under two evaluation metrics. We report our results on end-to-end spotting in the main manuscript, and additionally report the results under word spotting in Table \ref{tab:spot}.
\addtolength{\tabcolsep}{5pt} 
\begin{table}[!t]
\centering
\caption{Comparison with state-of-the-art scene text spotting techniques on \textbf{ICDAR2015}. `+oCLIP' refers to that the model are fine-tuned from the our pre-trained model on SynthText dataset. `S', `W', and `G' refer to word spotting with strong, weak, generic lexicon for ICDAR2015. `Full' refers to full lexicon for Total-Text.
}
\begin{tabular}{lccc}
\toprule
\textbf{Model}                    & \multicolumn{3}{c}{\textbf{ICDAR2015}}       \\ \cline{2-4} 
\textbf{}                         & \textbf{S}    & \textbf{W}    & \textbf{G}         \\
\midrule
TextDragon   \cite{feng2019textdragon}              & 86.2          & 81.6          & 68.0                   \\
\midrule
Mask TextSpotter-V3  \cite{liao2020mask}            & 83.1            & 79.1          & 75.1               \\
\textbf{Mask TextSpotter-V3+oCLIP}              &\textbf{84.1}, &\textbf{79.5} &\textbf{75.1}    \\  
\bottomrule
\end{tabular}
\label{tab:spot}
\end{table}

\addtolength{\tabcolsep}{-5pt}
\bibliographystyle{splncs04}
\bibliography{supp}